%% file: mainTex.tex
\documentclass[sigconf]{acmart}
\usepackage{import}
\usepackage{algorithmic}
\usepackage{algorithm}

\usepackage{textcomp}
\usepackage{multirow}
\usepackage{booktabs}
\usepackage{amsmath,xparse,mleftright}
\usepackage{balance}

\usepackage{enumitem}
\setitemize{noitemsep,topsep=0pt,parsep=0pt,partopsep=0pt}
\AtBeginDocument{%
  \providecommand\BibTeX{{%
    \normalfont B\kern-0.5em{\scshape i\kern-0.25em b}\kern-0.8em\TeX}}}

\copyrightyear{2022}
\acmYear{2022}
\setcopyright{acmcopyright}\acmConference[MM '22]{Proceedings of the 30th ACM
International Conference on Multimedia}{October 10--14, 2022}{Lisboa, Portugal}
\acmBooktitle{Proceedings of the 30th ACM International Conference on Multimedia
(MM '22), October 10--14, 2022, Lisboa, Portugal}
\acmPrice{15.00}
\acmDOI{10.1145/3503161.3548208}
\acmISBN{978-1-4503-9203-7/22/10}

\begin{document}

\title{CharFormer: A Glyph Fusion based Attentive Framework for High-precision Character Image Denoising}


\author{Daqian Shi}
\email{daqian.shi@unitn.it}
\orcid{0000-0003-2183-1957}
\affiliation{%
  \institution{College of Computer Science and Technology, Jilin University}
  \country{}
}
\additionalaffiliation{%
  \institution{DISI, University of Trento}
  \country{}
}

\author{Xiaolei Diao}
\email{xiaolei.diao@unitn.it}
\orcid{0000-0002-3269-8103}
\affiliation{%
  \institution{DISI, University of Trento}
  \country{}
}

\author{Lida Shi}
\email{shild21@mails.jlu.edu.cn}
\affiliation{%
  \institution{School of Artificial Intelligence, Jilin University}
  \country{}
}

\author{Hao Tang}
\email{hao.tang@vision.ee.ethz.ch}
\affiliation{%
  \institution{CVL, ETH Zurich}
  \country{}
}

\author{Yang Chi}
\email{yangchi19@mails.jlu.edu.cn}
\affiliation{%
  \institution{School of Artificial Intelligence, Jilin University}
  \country{}
}

\author{Chuntao Li}
\email{lct33@jlu.edu.cn}
\affiliation{%
  \institution{School of Archaeology, Jilin University}
  \country{}
}

\author{Hao Xu}
\email{xuhao@jlu.edu.cn}
\orcid{0000-0001-8474-0767}
\affiliation{%
  \institution{College of Computer Science and Technology, Jilin University}
  \country{}
}
\additionalaffiliation{%
  \institution{Symbol Computation and Knowledge Engineer of Ministry of Education, Jilin University}
  \country{}
}
\authornote{Corresponding author}


\renewcommand{\shortauthors}{Daqian Shi et al.}

\begin{abstract}
Degraded images commonly exist in the general sources of character images, leading to unsatisfactory character recognition results. Existing methods have dedicated efforts to restoring degraded character images. However, the denoising results obtained by these methods do not appear to improve character recognition performance. This is mainly because current methods only focus on pixel-level information and ignore critical features of a character, such as its glyph, resulting in character-glyph damage during the denoising process. In this paper, we introduce a novel generic framework based on glyph fusion and attention mechanisms, i.e., CharFormer, for precisely recovering character images without changing their inherent glyphs. Unlike existing frameworks, CharFormer introduces a parallel target task for capturing additional information and injecting it into the image denoising backbone, which will maintain the consistency of character glyphs during character image denoising. Moreover, we utilize attention-based networks for global-local feature interaction, which will help to deal with blind denoising and enhance denoising performance. We compare CharFormer with state-of-the-art methods on multiple datasets. The experimental results show the superiority of CharFormer quantitatively and qualitatively\footnote{Source code is available at \url{https://github.com/daqians/CharFormer}}.
\end{abstract}




\maketitle

\section{Introduction}
\import{sections/}{section1}

\section{Related Work}
\import{sections/}{section2}

\section{Proposed CharFormer}

\import{sections/}{section3}

\section{Experiments}
\import{sections/}{section4}

\section{Conclusion}
In this paper, we first discuss how the critical glyph information will affect the performance of character image denoising and further affect character recognition. Then we propose a novel glyph fusion-based attentive framework, i.e., CharFormer, for high-precision character image denoising. Our proposed framework focuses mainly on injecting specific information into the backbone model to improve the performance of image reconstruction. Thus, the attentive CharFormer blocks and the additional feature corrector module are designed to achieve this goal. Moreover, we introduce self-attention mechanisms into the CharFormer to build a powerful deep feature extractor that further improves the final denoising results. We compare CharFormer with state-of-the-art methods to evaluate its effectiveness, the experimental results show the superiority of CharFormer quantitatively and qualitatively.


\begin{acks}

This research is supported by the National Natural Science Foundation of China (62077027), the program of China Scholarships Council (No.202007820024), the Ministry of Science and Technology of the People's Republic of China (2018YFC2002500), and the Department of Science and Technology of Jilin Province, China (20200801002GH).
\end{acks}

\bibliographystyle{ACM-Reference-Format}
\balance
\bibliography{shi22}

\end{document}

%% file: sections/section1.tex
The optical character recognition (OCR) task \cite{chaudhuri2017optical} has been widely applied in industrial and academic applications in recent years. Clean character images are essential for improving the recognition performance of humans and machines \cite{5}. However, degradation exists commonly in general sources of character images, e.g., street view texts, handwritten manuscripts, and historical documents \cite{6}. Thus, learning a generic method to effectively denoise such character images becomes an attractive topic in the research community.

\begin{figure}[!t]
\setlength{\abovecaptionskip}{0pt}%
\setlength{\belowcaptionskip}{-5pt}%
	\centering
	\includegraphics[width=0.9\linewidth]{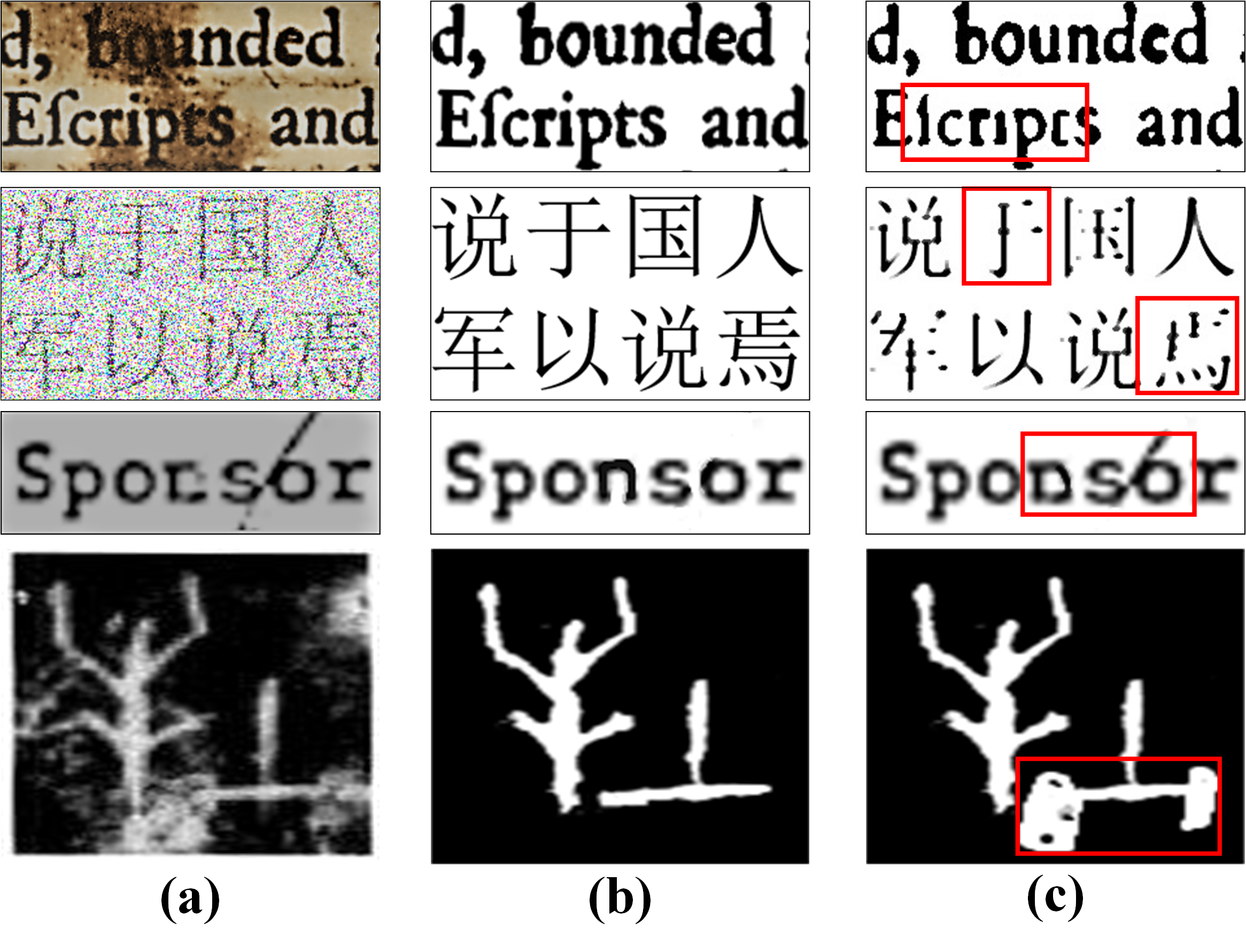}
    \caption{Character image denoising examples. (a) Character images with different degradation. (b) Correct denoising results. (c) Denoising results with incorrect glyphs, where the incorrect parts of glyphs are highlighted in red boxes. 
    \label{fig:1}}
\end{figure}

The glyph structure, which comes from the combination of character strokes, is the most crucial feature in maintaining the semantic coherence of the character \cite{7}. Characters are distinguished primarily by their glyphs rather than by other common features such as fonts \cite{8,11}. As a result, glyph information should be preserved properly in the low-level visual tasks of character images. Otherwise, the character will be unrecognized or incorrectly recognized because the semantic has changed. Figure~\ref{fig:1} demonstrates the character image denoising task, where (a) refers to images with various categories of noise, (b) and (c) refer to the results which are properly and incorrectly denoised, respectively. In Figure~\ref{fig:1}(c), we highlight the incorrect parts of the glyphs in red boxes. Compared to Figure~\ref{fig:1}(b), some character strokes of the former two images are incomplete due to the aggressive denoising strategy, e.g., the dot of the letter ``i'' is missing and the letter ``t'' is shaped closer to a ``c''. The noise in the latter two images is not precisely removed, where we can find that the letter ``o'' is still affected and is finally shaped closer to ``6'' (Figure~\ref{fig:1}(c) third row). In this paper, we focus on this low-level visibility degradation problem. Given a degraded character image,  we aim to precisely remove the diverse noise while not changing its inherent glyph, as shown in Figure~\ref{fig:1}(b).

Inspired by general image denoising, some character denoising methods focus on simulating different denoising scenarios via noise modeling. For instance, \cite{4} introduces a filter-based method for Chinese rubbing image restoration, which includes Gaussian and Wiener filtering. An adaptive method via the Gaussian mixture model is proposed for document noise removal \cite{3}. However, noise is formed and presented differently in diverse datasets (e.g., erosion noise and ink smear), making noise models more complex to simulate. As a result, these methods cannot achieve precise denoising results in practice. Excessive or insufficient noise removal will change the character glyph when dealing with such a complex degradation, as shown in Figure~\ref{fig:1}(c). To address this issue, some dedicated denoisers are designed for certain kinds of noise to obtain high-quality results \cite{6,10}. Deep generative networks are also introduced for achieving precise blind image restoration by image reconstruction \cite{1,2}. However, character images with damaged strokes still appear since these methods focus only on pixel-level image reconstruction, but do not consider glyph information.

To tackle the aforementioned issues, we propose a generic end-to-end framework based on glyph fusion and attention mechanisms (i.e., CharFormer) to achieve high-precision character image denoising. CharFormer introduces a parallel target task for capturing additional information and injecting it into the image-denoising backbone, which will maintain the consistency of character glyphs during denoising. Our framework consists of four modules (see Figure~\ref{fig:2}), including an input projector, a deep feature extractor, an output projector, and an additional feature corrector. The input projector aims to extract shallow features from the input images. The deep feature extractor is composed of novel CharFormer blocks (CFBs), each of which utilizes a residual self-attention block (RSAB) for global-local feature interaction. In addition, we propose a glyph structural network block (GSNB) to capture the glyph information that will be injected into RSABs. Both shallow and deep features are fed into the output projector to reconstruct high-precision clean-character images. Furthermore, the additional feature corrector intends to adjust the training of GSNBs for enhancing their glyph extraction ability.

Overall, our contributions can be summarized as follows:
\begin{itemize}[leftmargin=*]
    \item By introducing glyph information, we propose CharFormer, a novel attentive character image denoiser. It can effectively deal with various categories of noise in character images while maintaining the inherent glyph for character-semantic preservation.
    
    \item We propose an image reconstruction framework based on additional feature injection, where attention-based CharFormer blocks and the additional feature corrector module are designed for improving the quality of the output image by enhancing the corresponding feature.
    
    \item We compare CharFormer with state-of-the-art methods on datasets with different denoising difficulties. The experimental results show the superiority of CharFormer quantitatively and qualitatively, in particular on complex denoising tasks. 
\end{itemize}

%% file: sections/section2.tex
\subsection{Image Restoration}
As a low-level visual task, image restoration was originally solved by model-based methods \cite{12}. Various physics-based algorithms and filters \cite{Buades2005, Feng2019} are proposed for removing noise with certain models, e.g., Gaussian noise \cite{Gaussiannoise} or speckle noise \cite{bianco2018strategies}. Using 3D (inverse) transformations and shrinkage of the transform spectrum, BM3D is introduced for general image recovery \cite{BM3D-o}. In the last decade, learning-based methods, particularly based on convolutional neural networks (CNN), have gained in popularity in image restoration studies due to their impressive performance. Zhang et al. \cite{13} proposed a deep residual learning method named DnCNN to handle Gaussian denoising with an unknown noise level. General CNN-based architectures are also explored together with adversarial learning to restore degraded images \cite{14}, such as deblurring and dehazing. 
The attention mechanism has recently been applied in learning-based image restoration methods. For example, Zhang et al. \cite{15} proposed very deep residual channel attention networks (RCAN) to improve image super-resolution performance by filtering high-frequency information. Moreover, self-attention is also utilized in this field since the rise of visual transformers (ViT) \cite{16}. The image processing transformer (IPT) \cite{17} is introduced for three common low-level vision tasks, which is a pre-trained model using large-scale training data. To capture and maintain low-frequency visual features, Chen et al. \cite{18}and Wang et al. \cite{19} stacked self-attention blocks in a U-shaped network architecture to achieve effective image restoration.

\begin{figure*}[!t]
	\centering
	\includegraphics[width=0.95\linewidth]{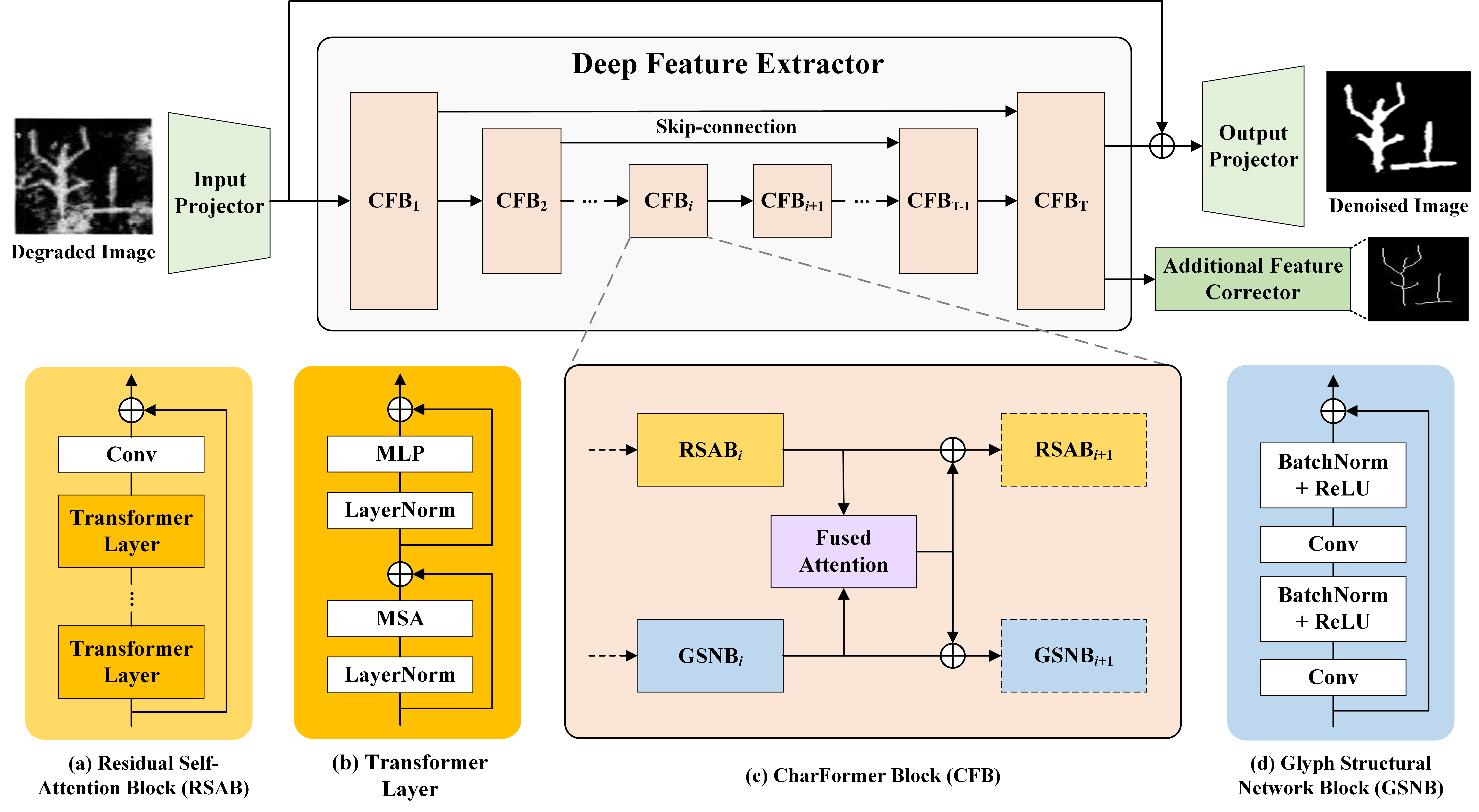}
    \caption{The overall architecture of the proposed CharFormer, where different blocks and layers are distinguished by colours. 
    \label{fig:2}}
\end{figure*}

\subsection{Character Image Denoising}
Character image denoising has become a growing topic due to the requirement for high-quality images from character recognition methods. Model-based denoising methods were first applied for such tasks. A comparative study \cite{4} applied several physics-based methods to denoise Chinese character rubbings, such as minimizing total variation \cite{chambolle2004algorithm} and wavelet transformation \cite{zhang2019wavelet}. However, such methods do not perform well in practice since character image degradation in the real world is more complex than synthetic noise. To address this issue, dedicated denoisers are designed for certain types of noise. For example, Gangamma et al. \cite{20} proposed an ensemble-based  denoising method to restore uneven backgrounds in degraded historical document images. A multi-spectral analysis-based method was introduced for recovering uneven backgrounds and ink smear in scanned historical documents \cite{9}.

Recently, learning-based denoising methods are starting to be applied to processing character images. A local adaptive threshold algorithm based on residual learning is proposed for dealing with non-uniformly illuminated, low contrast historic document images \cite{1}. Zhang et al. \cite{2} introduced a generative based method for Chinese calligraphic denoising by adding the combined Gaussian and salt-and-pepper noise. 
However, denoising results by the above-mentioned methods are still not satisfactory, since they lack considering the critical features, i.e., glyph information, which leads to the appearance of low-quality damaged characters and further affects character recognition and understanding. Thus, our objective is to solve these issues by introducing an attentive deep learning framework with glyph information injected.

%% file: sections/section3.tex
Given a degraded character image $I_{D} \in \mathbb{R}^{H\times W\times 3}$ with a spatial resolution of $H \times W$. Our goal is to reconstruct a high-precision noise-free character image $I_{R}$ accordingly. The intuition for applying glyph features in character image denoising is that the inherent glyph will help to maintain the structural consistency which is crucial for character recognition. As a result, we propose CharFormer, a new framework for additionally extracting and injecting such features into backbone networks to maintain the inherent glyph for the character image. Moreover, we introduce CFB based on transformers and fused attention to enhance fused feature extraction to improve the final image reconstruction performance. The following sections introduce the overall architecture and detail each component of CharFormer.

\subsection{Framework Architecture}
As shown in Figure~\ref{fig:2}, our proposed CharFormer consists of four modules: input projector, deep feature extractor, output projector, and additional feature corrector. 

\noindent \textbf{Input Projector.}
The input projector applies a $3 \times 3$ convolution layer with LeakyReLU, which aims to extract the shallow features of the input character image. The convolution layer is good at spatial feature extraction, which provides more stable early visual processing results \cite{21}. At the same time, shallow low-frequency features can be preserved for reconstructing the details in character images. Given a degraded image $I_{D}$, we have:
\begin{equation}
    F_{SF} = IP(I_{D}),
\label{equ:1}
\end{equation}
where $IP(\cdot)$ refers to the processing of input projector; $F_{SF}$ represents the shallow features, $F_{SF} \in \mathbb{R}^{H\times W\times C}$. Thus, the input image is mapped into a higher-dimensional feature space.

\begin{figure}[!t]
	\centering
	\includegraphics[width=1\linewidth]{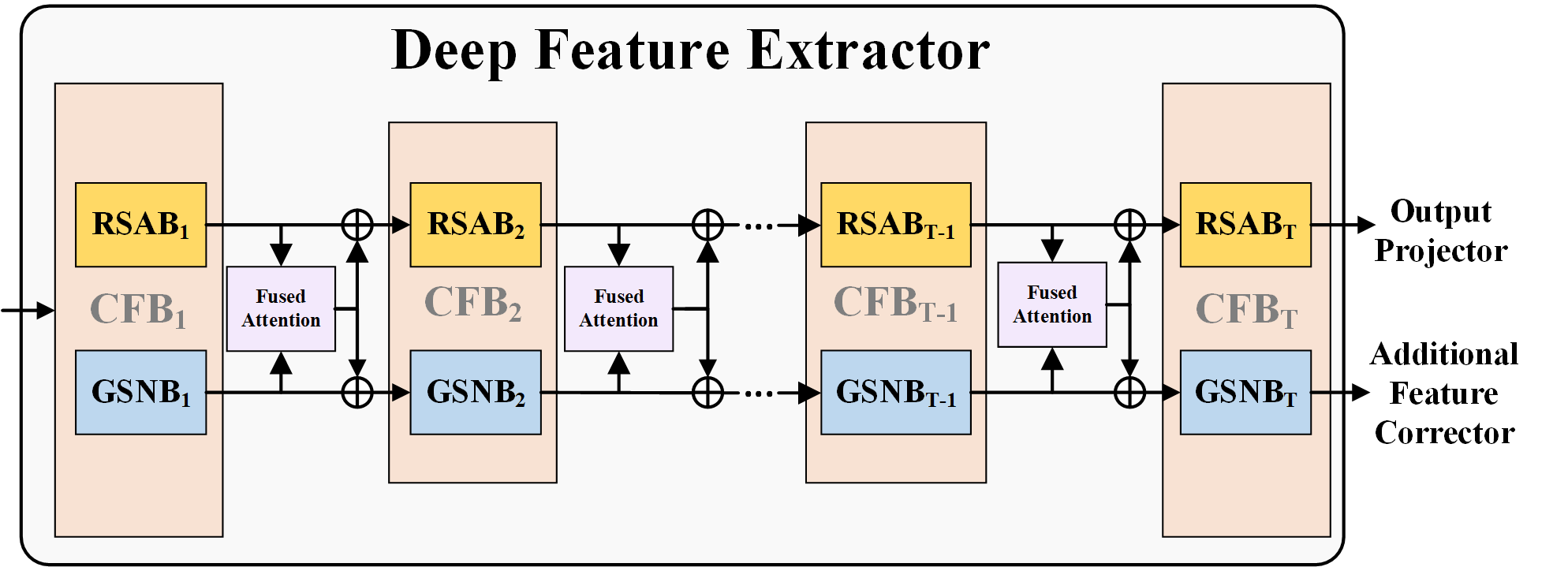}
    \caption{Details of CFB connections for demonstrating how each component of CFB works and the feature transformation procedure.  
    \label{fig:2.1}}
\end{figure}

\noindent \textbf{Deep Feature Extractor.}
Then we aim to extract deep features $F_{DF}$ from the shallow features $F_{SF}$. The deep feature extractor is developed into a U-shaped encoder-decoder structure, which is composed of $T$ CharFormer blocks (CFB). Thus, we extract $F_{DF} \in \mathbb{R}^{H\times W\times C}$ as:
\begin{equation}
    F_{DF} = DFE(F_{SF}),
\label{equ:1}
\end{equation}
where $DFE(\cdot)$ refers to the processing of deep feature extractor module. We apply skip connections between the CFBs in the encoder and decoder to transmit extraction results on different scales, which will help to capture spatial information and restore pixel-wise features \cite{22}. Particularly, the intermediate deep features $F_{DF_{i}}$ produced by CFBs are:
\begin{equation}
    F_{DF_{i+1}} = \left\{
    \begin{tabular}{ll}
    $CFB_{i+1}(F_{DF_i})$, & $0 \leqslant i \leqslant \frac{T}{2}$\\
    $CFB_{i+1}(F_{DF_i} + F_{DF_{T-i}})$,  & $\frac{T}{2} < i < T$ 
\end{tabular}\right.
\label{equ:3}
\end{equation}
where $CFB_{i}(\cdot)$ represents the $i$-th CharFormer block, $i = \{0,1,2...T\}$; $F_{DF_{i}}$ refers to the corresponding output of $CFB_{i}$; the concatenation operation will be applied before skip-connection. Note that each CFB parallelly organizes two components, RSAB and GSNB, to learn denoising information and glyph features, respectively, as Figure \ref{fig:2.1} shows. Thus, the intermediate feature $F_{DF_i}$ consists of $F_{REC_i}$ and $F_{GLY_i}$ that are output features of $i$-th RSAB and GSNB, respectively. The last CharFormer block $CFB_T$ will feed $F_{REC_T}$ to the output projector to reconstruct clean character images and $F_{GLY_T}$ to the additional feature extractor to adjust the training for glyph information extraction. 

\noindent \textbf{Output Projector.}
The output feature map of the deep feature extractor consists of two parts due to the specific structure of CFBs, where $F_{REC}$ will be fed into the output projector for clean character reconstruction. We obtain the final output $I_{R} \in \mathbb{R}^{H\times W\times C_i}$ as:
\begin{equation}
    I_{R} = OP(F_{SF} + F_{REC}),
\label{equ:4}
\end{equation}
where we input the aggregation of shallow features $F_{SF}$ and deep features $F_{REC}$ into the output projector $OP(\cdot)$ which is composed of three $3 \times 3$ convolution layers for rapid model convergence and precise pixel-wise image reconstruction. 

We utilize two common loss functions to better illustrate the effectiveness of our proposed framework. For pixel level image reconstruction, we introduce the pixel loss for the reconstructed character image $I_R$ as:
\begin{equation}
    \mathcal{L}_1(I_R) = \left \| I_{GT} - I_R  \right \|_1,
\label{equ:5}
\end{equation}
where $\mathcal{L}_1(\cdot)$ represents the pixel loss function and $I_{GT}$ refers to the ground-truth noise-free character image. The perceptual loss $\mathcal{L}_P(\cdot)$ \cite{24} is proposed by considering the feature level information comparison and the global discrepancy. Based on a VGG16 model $VGG(\cdot)$ pretrained on the ImageNet dataset, we define perceptual loss for $I_R$ as:
\begin{equation}
    \mathcal{L}_P(I_R) = \left \| VGG(I_{GT}) - VGG(I_R)  \right \|_1.
\label{equ:6}
\end{equation}


\noindent \textbf{Additional Feature Corrector.}
We introduce CharFormer and its critical component CFBs to extract and inject specific features during image reconstruction, where the additional feature corrector aims to adjust the feature extraction. For our character image denoising task, we consider the glyph information as such specific features to improve the quality of denoised characters. When dealing with character images, skeletonization methods are widely applied to present glyphs and structural strokes of characters, e.g., English, Chinese and Korean characters \cite{7,ko2021skelgan}. 

Thus, in the additional feature corrector, we utilize an existing method \cite{23} based on mathematical morphology, which is particularly designed for character skeletonization. By inputting the ground truth of the noisy character image $I_{GT}$, we obtain the skeletonized binary image $I_{GT_S} \in \mathbb{R}^{H\times W}$. We utilize the same loss functions in this module with the output projector, thus, we have $\mathcal{L}_1(I_S) =  \left \| I_{GT_S} - I_S  \right \|_1$ and $\mathcal{L}_P(I_S) = \left \| VGG(I_{GT_S}) - VGG(I_S)  \right \|_1$. Finally, we define the overall loss function for CharFormer as: 
\begin{equation}
    \mathcal{L} = \mathcal{L}_1(I_R) + \mathcal{L}_P(I_R) + \phi (\mathcal{L}_1(I_S) + \mathcal{L}_P(I_S)),
\label{equ:7}
\end{equation}
where $\phi$ refers to the weight of the corresponding loss functions\footnote{We set the weight of glyph losses $\phi = 1$ in the following experiments.}.

\subsection{CharFormer Block}
We introduce CharFormer Block as the critical component for injecting glyph information into the backbone of the deep feature extractor. Each CFB consists of a residual self-attention block (RSAB) as the denoising backbone, a glyph structural network block (GSNB) for glyph information extraction, and a fusion attention layer for feature injection, as shown in Figure~\ref{fig:2}(c) and Figure~\ref{fig:2.1}.
 
\noindent \textbf{Residual Self-Attention Block.}
Figure~\ref{fig:2}(a) shows the structure of RSAB, which is a residual block with $K$ transformer layers and a convolutional layer. We aim to take advantage of the self-attention mechanism for capturing long-range spatial dependencies and apply a regular convolution layer for improving the translational equivariance of the network. Moreover, the skip connection provides a path for transmitting aggregated multi-scale features to the output projector. 

RSABs are able to output feature maps on different scales since CFBs are stacked as a U-shaped encoder-decoder. We apply down-sampling RSABs in the encoder, specifically, given an input feature map $F_{REC_0} \in \mathbb{R}^{H\times W\times C}$, the $i$-th RSAB will output $F_{RSAB_{i}} \in \mathbb{R}^{\frac{H}{2^i}\times \frac{W}{2^i}\times 2^iC}$, $ i\leqslant \frac{T}{2}$. For feature reconstruction, we introduce transposed convolution in upsampling RSABs to recover the size of feature maps.

Inspired by \cite{16,18}, we exploit the self-attention mechanism for visual feature extraction by using a standard transformer. As shown in Figure~\ref{fig:2}(b), each transformer layer applies multi-head self-attention MSA($\cdot$) and multi-layer perceptron MLP($\cdot$). Thus, the output feature of the $l$-th transformer layer in RSAB is defined as:
\begin{equation}
\begin{aligned}
    F^{\prime}_l &= {\rm MSA}({\rm LN}(F_{l-1})) + F_{l-1}, \\
    F_l &= {\rm MLP}({\rm LN}(F^{\prime}_{l})) + F^{\prime}_{l},
\end{aligned}
\label{equ:8}
\end{equation}
where LN($\cdot$) refers to the layer normalization operation; $F^{\prime}_l$ and $F_l$ denote the intermediate and final output features of the transformer layer, respectively. For applying MSA, a feature map $\hat{X} \in \mathbb{R}^{H\times W\times C}$ will be separated into non-overlapping patches with the size of $M \times M$. Given a patch $X \in \mathbb{R}^{M^2 \times C}$, the calculation of self-attention in MSA can be formulated as: 
\begin{equation}
    Attention(Q,K,V) = {\rm Softmax}(\frac{QK^T}{\sqrt{d}}+B)V,
\label{equ:9}
\end{equation}
where query, key, and value matrices $Q,K,V$ are calculated by their corresponding projection matrices with the patch $X$; $B$ represents learnable relative position encoding \cite{25}. 

\noindent \textbf{Glyph Structural Network Block.}
GSNB is designed to extract the glyph information that will be injected into the backbone network RSABs, it stacks $3 \times 3$ convolution layers in a residual block\footnote{The depth of each RSAB and GSNB are set as 3 and 2, respectively.}, as shown in Figure~\ref{fig:2}(d). Similar to RSABs, GSNBs also include down-sampling and up-sampling cases since they need to keep the same scale as the corresponding RSABs. Given an input glyph feature $F_{GLY_0} \in \mathbb{R}^{H\times W\times C}$, the GSNB in $i$-th CFB will output $F_{GSNB_{i}}$ which has the same size to $F_{RSAB_{i}}$.

\noindent \textbf{Fused Attention.} 
Before injecting the feature $F_{GSNB}$ to the backbone network, a fused attention layer is applied to weigh the key information that will affect the final denoising. Inspired by \cite{26}, channel attention is applied for providing effective tokens from selective feature maps. We also use spatial attention to enhance dependency extraction in the spatial axes. Given an input feature $F_k$ , the output $FA(F_k)$ of the fused attention layer is: 
\begin{equation}
    FA(F_k) = M_s(M_c(F_k)+F_k)+M_c(F_k)+F_k,
\label{equ:10}
\end{equation}
where $M_c,M_s$ are the channel and spatial attention, respectively. 

Thus, we formulate the reconstruction feature $F_{REC}$ for the $i$-th CFB as:
\begin{equation}
    F_{REC_i} = FA(F_{RSAB_i}+F_{GSNB_i})+F_{RSAB_i},
\label{equ:11}
\end{equation}
Correspondingly, the additional glyph feature $F_{GLY}$ is formulated as:
\begin{equation}
    F_{GLY_i} = FA(F_{RSAB_i}+F_{GSNB_i})+F_{GSNB_i},
\label{equ:12}
\end{equation}
where $F_{REC_i}$ and $F_{GLY_i}$ will be fed into RSAB and GSNB in the next CFB, respectively. Note that $F_{REC_0} = F_{GLY_0} = F_{SF}$.

%% file: sections/section4.tex
\begin{table*}[!t]
\centering
\caption{Quantitative comparisons (average PSNR, SSIM, and character recognition accuracy) with state-of-the-art methods on four datasets. The best and second-best results are highlighted in {\color[HTML]{FF0000} red} and {\color[HTML]{0070C0} blue} colors, respectively.}
\label{tab:1}
\resizebox{17.7cm}{!}{

\begin{tabular}{@{}lcccccccccccc@{}}
\toprule
\multirow{2}{*}{Method} & \multicolumn{3}{c}{$Dataset1$}                                         & \multicolumn{3}{c}{$Dataset2$}                     & \multicolumn{3}{c}{$Dataset3$}                     & \multicolumn{3}{c}{$Dataset4$}                      \\ \cmidrule(l){2-4}  \cmidrule(l){5-7} \cmidrule(l){8-10}  \cmidrule(l){11-13}
                         & PSNR$\uparrow$ & SSIM$\uparrow$ & \multicolumn{1}{c}{Acc$_R$$\uparrow$} & PSNR$\uparrow$ & SSIM$\uparrow$ & Acc$_R$$\uparrow$ & PSNR$\uparrow$ & SSIM$\uparrow$ & Acc$_R$$\uparrow$ & PSNR$\uparrow$ & SSIM$\uparrow$  & Acc$_R$$\uparrow$ \\ \midrule
Raw Image                & 16.33                        & 0.7978                        & 0.6931                        & 13.03                        & 0.2852                        & -                                & 12.30                        & 0.1049                        & 0.5689                                & 8.790                         & 0.3969                        & 0.4300                        \\
IPT \cite{17}                     & 23.72                        & {\color[HTML]{FF0000} 0.9027} & 0.8560                         & {\color[HTML]{0070C0} 21.04} & 0.8293                        & {\color[HTML]{0070C0} 0.6854}         & 22.60                        & 0.8897                        & 0.8144                                & {\color[HTML]{0070C0} 18.95} & {\color[HTML]{0070C0} 0.8663} & {\color[HTML]{0070C0} 0.5967} \\
DnCNN \cite{13}                   & 21.04                        & 0.8763                        & 0.7323                        & 19.55                        & 0.7978                        & 0.3149                                & 22.35                        & 0.8795                        & 0.7922                                & 15.64                        & 0.7053                        & 0.4611                        \\
CIDG \cite{2}                    & 21.88                        & 0.8871                        & 0.7559                        & 20.65                        & {\color[HTML]{0070C0} 0.8623} & 0.2471                                & 21.04                        & 0.8611                        & 0.7689                                & 17.71                        & 0.7922                        & 0.5533                        \\
UFormer \cite{19}                 & 23.86                        &  0.8970 & 0.8326                        & 21.01                        & 0.8221                        & 0.6693                                & {\color[HTML]{FF0000} 22.83} & {\color[HTML]{0070C0} 0.8909} & 0.8267                                & 18.90                        & 0.8537                        & 0.5889                        \\
InvDN \cite{InvDN}                   & 22.40                        & 0.8807                        & 0.8374                        & 20.49                        & 0.8077                        & 0.5917                                & 22.16                        & 0.8726                        & 0.8044                                & 18.35                        & 0.8332                        & 0.5722                        \\
BM3D \cite{BM3D}                & 19.83 & 0.8424 & 0.7145 & 20.67  & 0.8607  & 0.6122    & 22.58   & 0.8837    &  0.8163  & 12.74  & 0.5961 & 0.4905 \\ 

VDN \cite{guo2019toward}        & 23.66 & 0.8857 & 0.8319 & 20.97  & 0.8470  & 0.6409    & 22.63  & 0.8907    &  0.8897  & 18.43  & 0.8331 & 0.6016 \\ 

TransUNet \cite{18}                & {\color[HTML]{0070C0} 23.92} & {\color[HTML]{0070C0} 0.8998} & {\color[HTML]{FF0000} 0.8621} & 20.83                        & 0.8592                        & 0.5579                                & 22.47                        & 0.8732                        & {\color[HTML]{0070C0} 0.8356}         & 17.98                        & 0.8136                        & 0.5800                        \\ \midrule

CharFormer (Ours)        & {\color[HTML]{FF0000} 24.08} & 0.8985                        & {\color[HTML]{0070C0} 0.8553} & {\color[HTML]{FF0000} 21.07} & {\color[HTML]{FF0000} 0.8637} & {\color[HTML]{FF0000} 0.7259}         & {\color[HTML]{0070C0} 22.71} & {\color[HTML]{FF0000} 0.8913} & {\color[HTML]{FF0000} 0.8433}         & {\color[HTML]{FF0000} 19.87} & {\color[HTML]{FF0000} 0.8772} & {\color[HTML]{FF0000} 0.6856} \\ \bottomrule

\end{tabular}}
\end{table*}

\begin{figure*}[!t]
\setlength{\abovecaptionskip}{2pt}%
\setlength{\belowcaptionskip}{0pt}%
	\centering
	\includegraphics[width=0.95\linewidth]{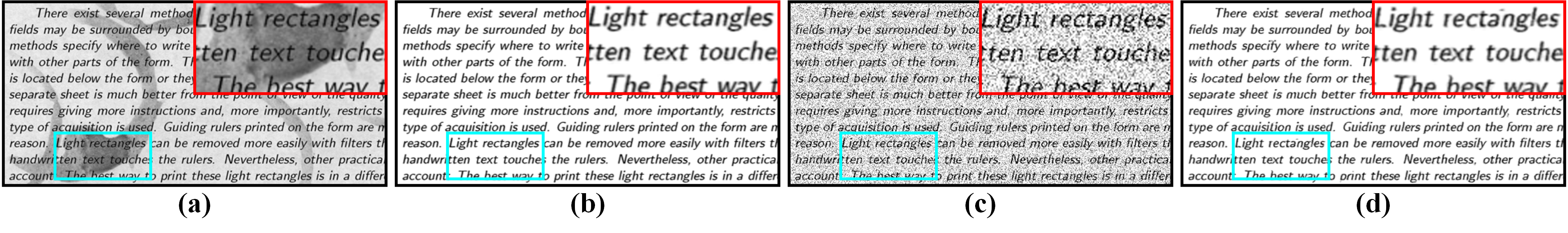}
    \caption{Visual results of CharFormer on printed English document images. (a)-(b) raw images and results of $Dataset1$. (c)-(d) raw images and results of $Dataset2$.
    \label{fig:3}}
\end{figure*}

\begin{figure}[!t]
\setlength{\abovecaptionskip}{2pt}%
\setlength{\belowcaptionskip}{0pt}%
	\centering
	\includegraphics[width=0.95\linewidth]{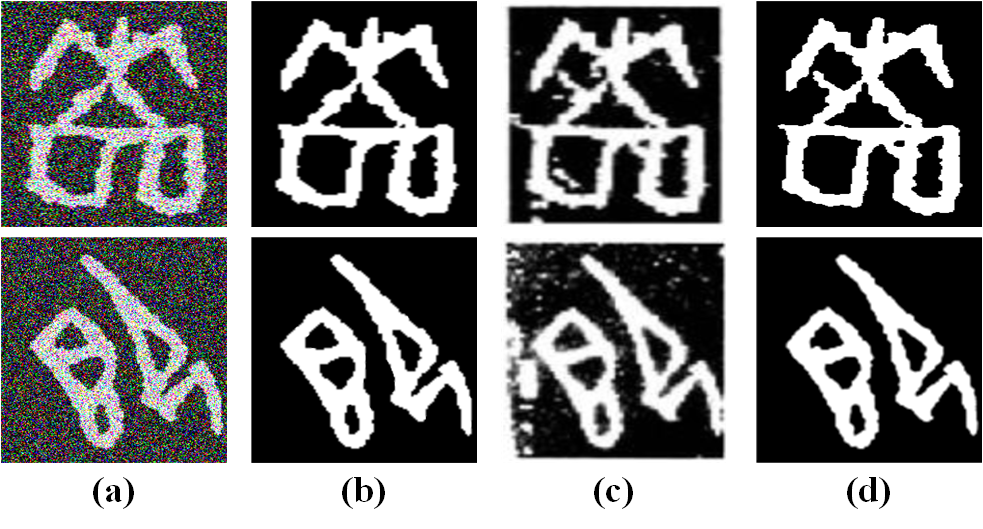}
    \caption{Visual results of CharFormer on historical Chinese character images. (a)-(b) raw images and results of $Dataset3$. (c)-(d) raw images and results of $Dataset4$.
    \label{fig:4}}
\end{figure}

\subsection{Experimental Setups}
\noindent\textbf{Datasets.}
Our experiments utilize different datasets to present a comprehensive validation. We consider two sources, which cover document-level and character-level data in two languages, to provide data for model training and testing. We select the document-level image set \textit{denoising-dirty-documents}\footnote{https://www.kaggle.com/c/denoising-dirty-documents/data} which contains printed English words in 18 different fonts. Thus, we obtain $Dataset1$ which involves noisy raw images with uneven backgrounds. We also generate $Dataset2$ by adding mixed Gaussian and speckle noise (noise variance $\sigma$ = 5) on the ground truth of document-level images.

In addition, we introduce a character-level image set \textit{noisy-character-images}\footnote{https://github.com/daqians/Noisy-character-image-benchmark.} that collects handwritten historical Chinese characters with natural noise. $Dataset3$ simulates a blind denoising scenario by randomly adding mixed Gaussian and speckle noise (noise variance $\sigma = [10,50]$) to the ground truth of these character-level images, where the ground truth images are manually annotated by five philologists. At last, we define the noisy raw character images as $Dataset4$ to provide cases with complex noise. It is worth noting that the denoising difficulty for these four datasets gradually increases as the complexity of these noise models increase.

\noindent\textbf{Baseline Methods.}
We compare CharFormer with state-of-the-art image denoising methods, including calligraphic image denoising GAN (CIDG) \cite{2}, DnCNN \cite{13}, InvDN \cite{InvDN}, VDN \cite{guo2019toward}, some transformer-based methods (UFormer \cite{19}, IPT \cite{17}, and TransUNet \cite{18}) and a classic denoising algorithm BD3M \cite{BM3D}. Need to notice that all the methods we applied are proposed for character image denoising or widely applied in general denoising scenarios. To fairly compare these methods, we provide the same training environment and data for all the above methods.

\subsection{Quantitative Evaluation}
Table~\ref{tab:1} demonstrates quantitative comparisons between CharFormer and state-of-the-art image denoising methods on the four datasets. Two commonly applied metrics of low-level vision tasks are introduced to evaluate pixel-level denoising performance, i.e., peak signal-to-noise ratio (PSNR) and the structural similarity index measure (SSIM) \cite{PSNR}. Higher PSNR and SSIM values indicate that the denoised image is closer to the target image under the pixel-wised comparison. In addition, we also introduce a metric for character-level evaluation, i.e., optical character recognition accuracy (Acc$_R$), to validate if the denoising results effectively improve character recognition performance compared to raw images. We exploit two publicly accessible OCR tools\footnote{https://www.ocr2edit.com/\label{ocr1}}$^,$\footnote{http://api.shufashibie.com/page/index.html\label{ocr2}} to recognize English words ($Dataset1 \& 2$) and historical Chinese ($Dataset3 \& 4$), respectively. The metrics on raw images are involved in the table to visually present the improvements.

The results indicate that CharFormer surpasses all state-of-the-art methods on $Dataset2$ and $Dataset4$, and obtains promising performance on $Dataset1$ and $Dataset3$. For document-level datasets, most of the comparison methods perform well since the uneven background does not visibly degrade the raw images in $Dataset1$. CharFormer leads the results in $Dataset2$, which indicates the effectiveness of the proposed model. Some attention-based methods also show their competitiveness, such as IPT and CIDG. Note that our good performance on $Dataset1 \& 2$ proves that our method is adaptive to various fonts of characters since the character in different fonts still share a similar glyph. At the same time, character recognition results also show visible differences between $Dataset1$ and $Dataset2$, mainly because characters in document images are easily affected by complex degradation due to their smaller size. 

For character-level datasets, some methods perform better in $Dataset3$ because they are designed for blind denoising, such as InvDN and DnCNN. Some transformer-based methods like Uformer and TransUNet also achieve competitive performance. When testing CharFormer on $Dataset4$, the performance further increases by a large margin on three metrics, achieving better performance than transformer-based methods. Methods like DnCNN and CIDG poorly perform on $Dataset4$ since they are designed based on synthetic noise rather than complex real-world noise. BM3D achieves limited results on $Dataset1 \& 4$ but better on $Dataset2 \& 3$ since it is not designed for large-scale noise, such as uneven backgrounds. Overall, we find that CharFormer is outstanding in all character denoising cases, especially for complex denoising cases like $Dataset2, 3 \& 4$ which contain mixed noise and real-world noise. 

\subsection{Qualitative Evaluation}
\begin{figure*}[hpt]
	\centering
	\setlength{\abovecaptionskip}{5pt}%
    \setlength{\belowcaptionskip}{0pt}%
	\includegraphics[width=0.92\linewidth]{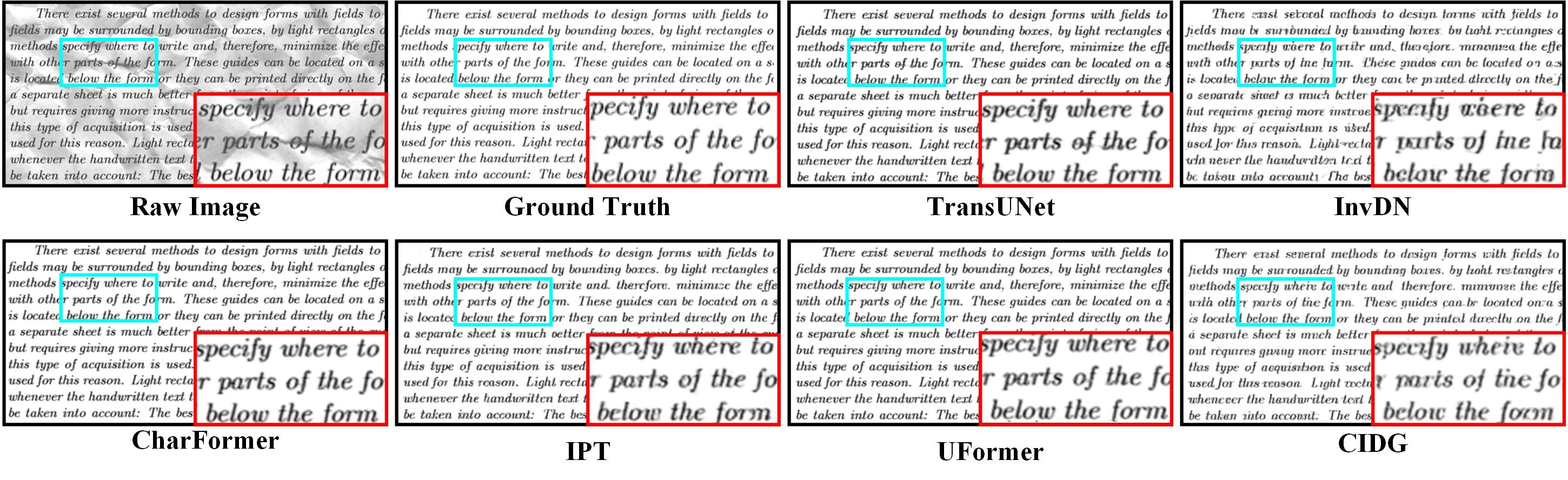}
    \caption{Qualitative comparisons on $Dataset1$ which contains uneven background noise.  
    \label{fig:a1}}
\end{figure*}

\begin{figure*}[hpt]
	\centering
	\setlength{\abovecaptionskip}{5pt}%
    \setlength{\belowcaptionskip}{0pt}%
	\includegraphics[width=0.92\linewidth]{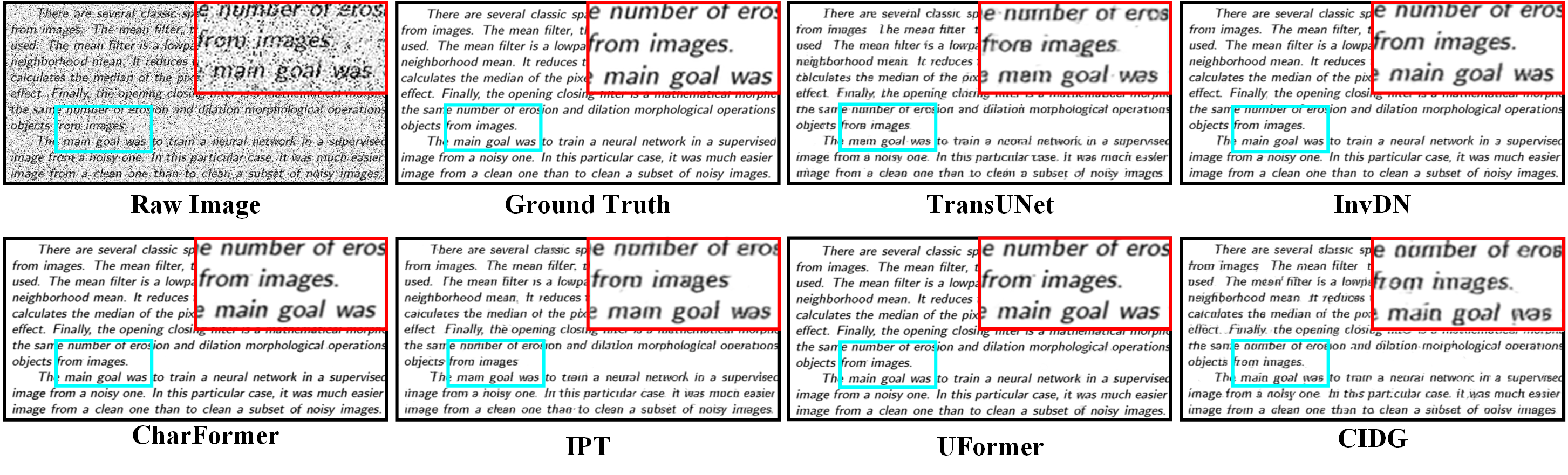}
    \caption{Qualitative comparisons on $Dataset2$ which contains mixed Gaussian and speckle noise.  
    \label{fig:a2}}
\end{figure*}

\begin{figure*}[hpt]
	\centering
	\setlength{\abovecaptionskip}{5pt}%
    \setlength{\belowcaptionskip}{0pt}%
	\includegraphics[width=0.92\linewidth]{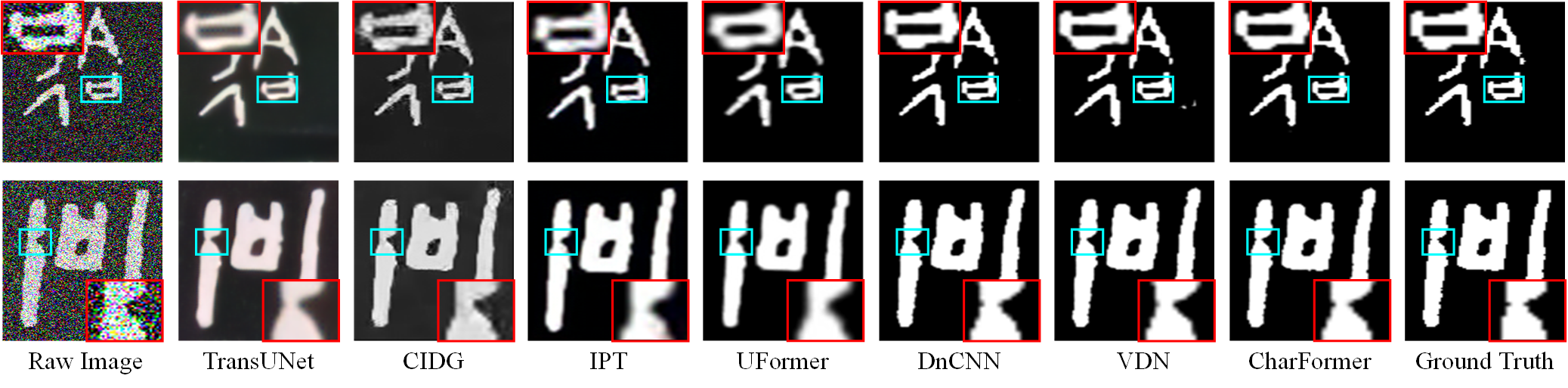}
    \caption{Qualitative comparisons on $Dataset3$ which contains random level Gaussian noise.  
    \label{fig:a3}}
\end{figure*}

\begin{figure*}[hpt]
	\centering
	\setlength{\abovecaptionskip}{3pt}%
    \setlength{\belowcaptionskip}{0pt}%
	\includegraphics[width=0.92\linewidth]{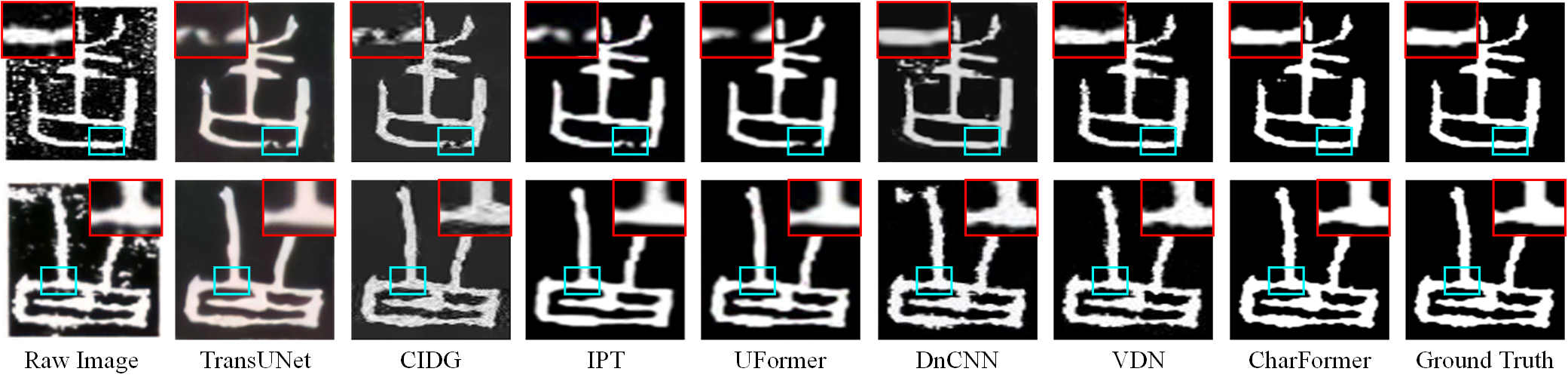}
    \caption{Qualitative comparisons on $Dataset4$ which contains complex real-world noise.  
    \label{fig:a4}}
\end{figure*}

Correspondingly, we provide visual results of CharFormer on the four datasets. As shown in Figure~\ref{fig:3}(b) and (d), CharFormer can effectively remove the uneven background and mixed Gaussian and speckle noise in printed English document images. We zoom in to demonstrate the denoising details where the character can be recognized and the glyph is not damaged. Figure~\ref{fig:4} presents the denoising results on historical Chinese character images, where CharFormer shows great effectiveness in dealing with different kinds of noise in Figure~\ref{fig:4}(b) and (d). More specifically, wind erosion and broken edge noise are properly restored in the two cases in Figure~\ref{fig:4}(c). In blind denoising cases Figure~\ref{fig:4}(a), random noise-level Gaussian noise has been removed without affecting the character due to the use of glyph information. Figure~\ref{fig:a1} - \ref{fig:a4} presents qualitative comparisons with state-of-the-art methods on four datasets. We find that the results of CharFormer outperform others in most cases, since the inherent character glyphs are preserved while precisely removing noise. Moreover, the powerful U-shaped feature extractor built by the proposed CFBs is another main reason for achieving good denoising performance. 

\subsection{Ablation Study}
Ablation studies are used to provide detailed proof that each design in CharFormer is reasonable and effective.

\begin{figure}[!t]
\setlength{\abovecaptionskip}{5pt}%
\setlength{\belowcaptionskip}{0pt}%
	\centering
	\includegraphics[width=0.95\linewidth]{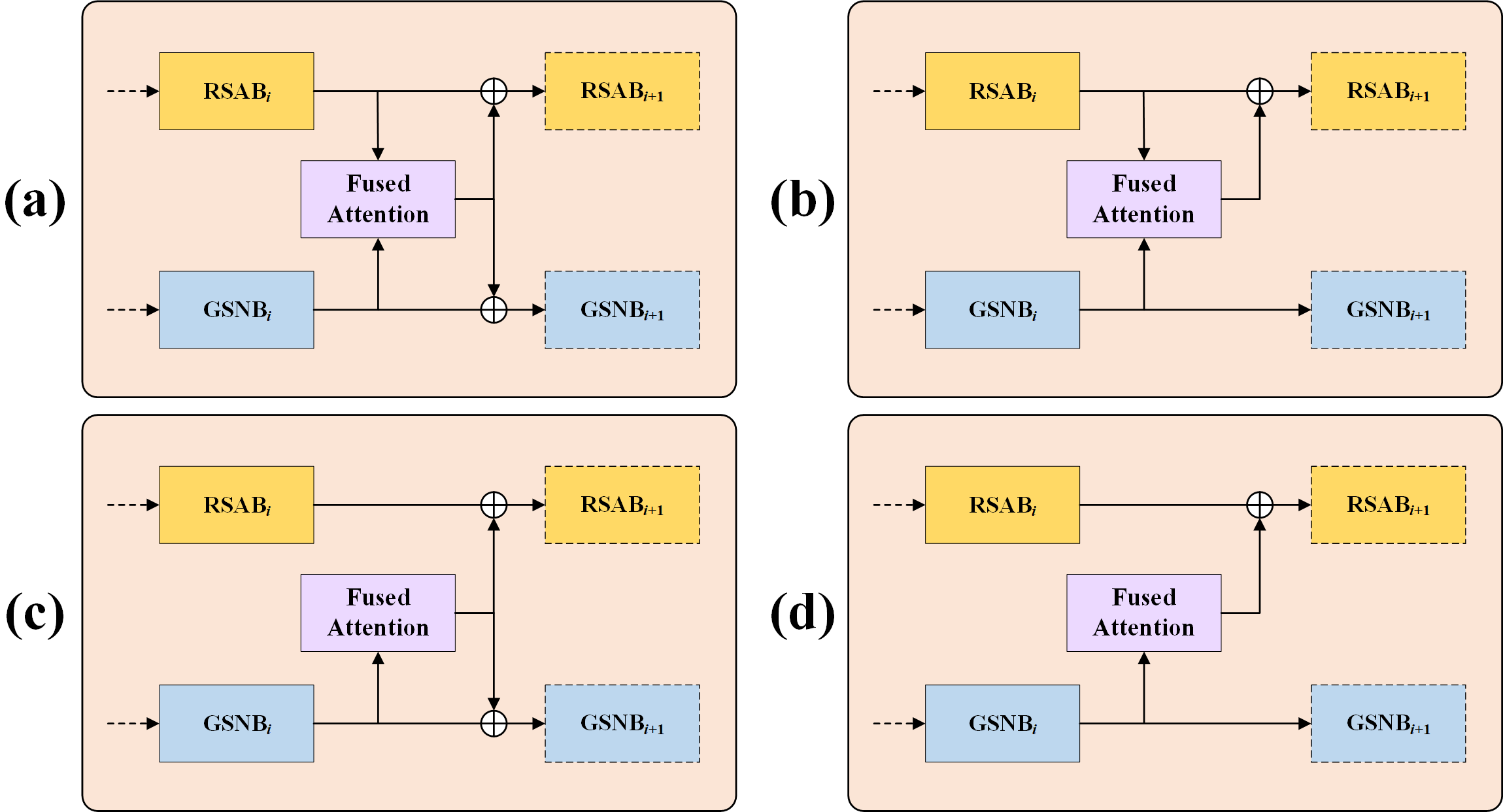}
    \caption{Four connection schemes for injecting glyph information.}
    \label{fig:6}
        \vspace{-0.3cm}
\end{figure}

\begin{figure}[!t]
\setlength{\abovecaptionskip}{5pt}%
\setlength{\belowcaptionskip}{0pt}%
	\centering
	\includegraphics[width=0.95\linewidth]{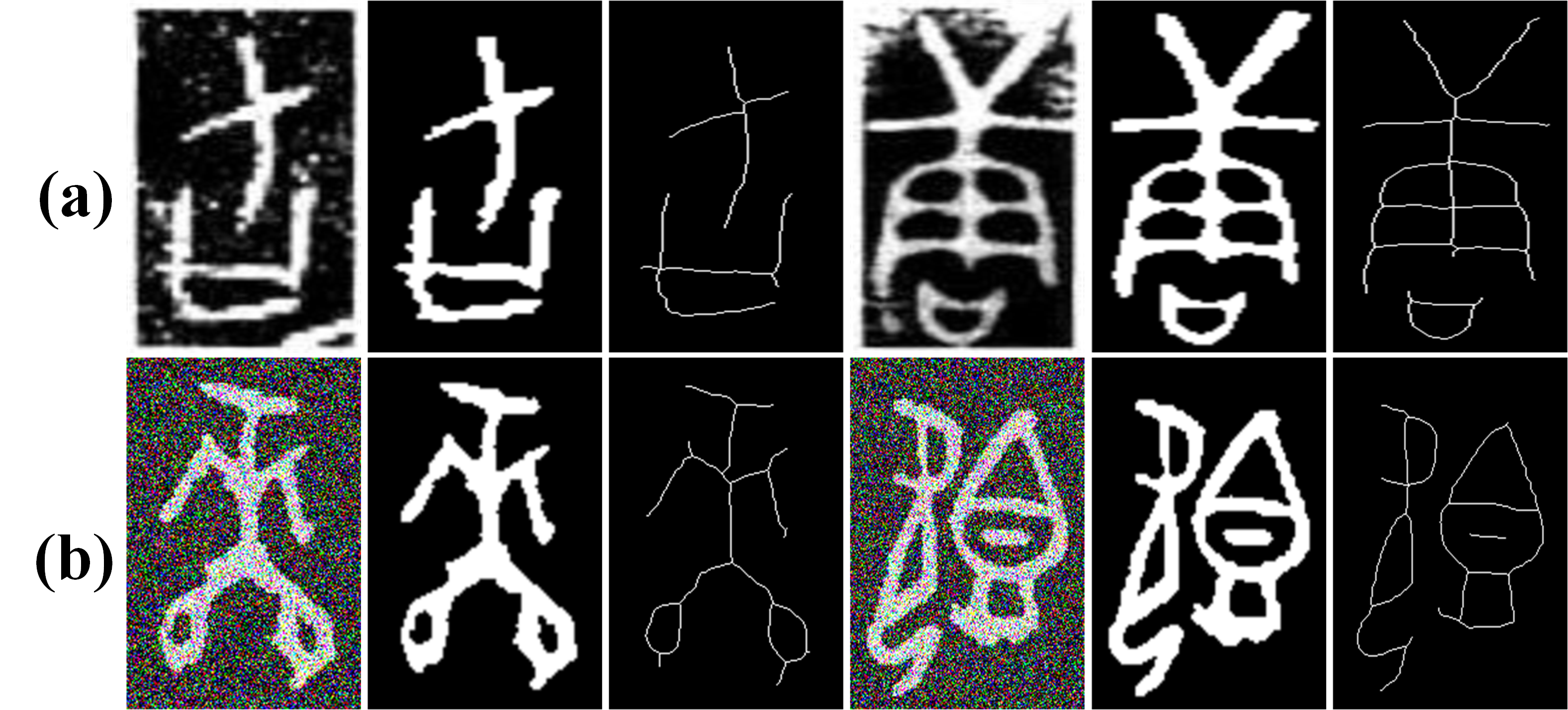}
    \caption{Validation of glyph information extraction ability for character images in $Dataset3$ and $Dataset4$.}
    \label{fig:5}
    \vspace{-0.3cm}
\end{figure}

\noindent\textbf{Impact of Fused Attention and Connection Schemes.}
In this paper, we propose fused attentive CharFormer blocks as the core design for information selection and additional feature injection. To validate the impact of fused attention, we apply a backbone that directly connects the output of $GSNB_i$ with $RSAB_i$, and four different connection schemes (Con$_a$...Con$_d$) as shown in Figure~\ref{fig:6}. CFBs are currently applying the connection scheme in Figure~\ref{fig:6}(a). 
The results on $Dataset4$ can be found in Table~\ref{tab:2}. Compared with the results of the backbone model, PSNR/SSIM both increase by a large margin with applying fused attention in four connection schemes. By applying the dual-residual connection, Figure~\ref{fig:6}(a) achieves the best performance, which validates our current design of CFBs.

\begin{table}[!t]
\centering
\captionsetup{font=small, labelfont=bf}
\setlength{\abovecaptionskip}{0pt}%
\setlength{\belowcaptionskip}{2pt}%
\caption{Ablation study results on connection schemes.}
\label{tab:2}
\begin{tabular}{@{}cccccc@{}}
\toprule
     & Backbone & Con$_a$ & Con$_b$ & Con$_c$ & Con$_d$ \\ \midrule
PSNR $\uparrow$ & 19.32    & {\color[HTML]{FF0000} 19.87} & 19.8   & 19.57  & 19.65  \\
SSIM $\uparrow$ & 0.8577   & {\color[HTML]{FF0000} 0.8772} & 0.8708 & 0.8649 & 0.8683 \\ \bottomrule
\end{tabular}
\end{table}

\noindent\textbf{Validation of Glyph Information Extraction.}
To prove that the GSNBs can effectively capture glyph information from noisy character images and inject it into the denoising backbone, we visualize the output of the additional feature corrector. As shown in Figure \ref{fig:5}, row (a) and (b) refers to the generated glyph visualization for images in $Dataset4$ and $Dataset3$, respectively. For each case in Figure \ref{fig:5}, the former two images refer to the noisy character image and its ground truth and the last image is the character skeleton, where we can find that the glyphs are extracted properly.

\noindent\textbf{Impact of using Transformers in RSAB.}
We apply the self-attention mechanism and transformers in CFB to obtain a more powerful deep feature extractor. In this experiment, we use the convolution layer instead of the transformer layer in RSAB as the backbone model, RSAB with 1 (B+1TLs) or 3 (B+3TLs) transformer layers as comparison cases, Table~\ref{tab:3} shows the results of testing on $Dataset4$. It indicates that the CFBs perform better with using self-attention mechanisms in RSABs. Thus, we set $l = 3$ as the number of Transformer layers in each RSAB.

\begin{table}[!t]
\centering
\captionsetup{font=small, labelfont=bf}
\setlength{\abovecaptionskip}{2pt}%
\setlength{\belowcaptionskip}{2pt}%
\caption{Ablation study results on RSAB.}
\label{tab:3}
\begin{tabular}{@{}cccc@{}}
\toprule
     & Backbone & B+1TLs & B+3TLs \\ \midrule
PSNR/SSIM $\uparrow$ & 18.94/0.8525 & 19.31/0.8603 & {\color[HTML]{FF0000} 19.87/0.8772} \\ \bottomrule
\end{tabular}
\end{table}


\subsection{User study}
We have discussed character-level improvements in character images by comparing the OCR accuracy of denoised characters provided by CharFormer. For comprehensively validating the character-level improvement, we also provide explicit user study on denoised character images. Three metrics are introduced for evaluating the raw and denoised character images, which are (1) the \textbf{Clarity} of the text; (2) the \textbf{Integrity} of the glyph; (3) the \textbf{Interference} of the remaining noise. Note that all three metrics are rated from 1 to 10 and higher ratings indicate better image quality. We provide 100 images (randomly selected from $Dataset1-4$) to 12 linguistics students. Table \ref{tab:4} shows the average rating of raw and denoised character images by CharFormer. We can find that the rating on three metrics significantly increases after denoising, which further indicates that our method can also benefit human character recognition.

\begin{table}[!t]
\centering
\captionsetup{font=small, labelfont=bf}
\setlength{\abovecaptionskip}{2pt}%
\setlength{\belowcaptionskip}{2pt}%
\caption{User study results.}
\label{tab:4}
\begin{tabular}{@{}cccc@{}}
\toprule
     & Clarity $\uparrow$ & Integrity $\uparrow$ & Interference-free $\uparrow$ \\ \midrule
Raw/Denoised & 4.72/8.90 & 7.79/9.35 & 5.12/9.63 \\ \bottomrule
\end{tabular}
\end{table}